\documentclass[12pt, a4]{article}
\usepackage[T1]{fontenc}
\usepackage{graphicx}
\usepackage{amsmath}
\usepackage{amssymb}
\usepackage{smartdiagram}
\usepackage{hyperref}

\def\be{\begin{equation}}
\def\ee{\end{equation}}
\def\ff#1{{\mbox{\boldmath{$#1$}}}}

\def\x{\ff{x}}
\def\mat#1{{\left( \begin{array}{cccc}#1\end{array} \right)}}

\def\w{\omega}
\def\E#1{\mathbb{E}\left[ #1 \right]}

\begin{document}

\title{Analysis of Parameter Settings for the Bat Algorithm Using Variance Evolution}

\author{Xin-She Yang\thanks{Corresponding Author}, \;\;\;\;\;  Mehmet Karamanoglu \\[10pt]
Faculty of Science and Technology,  Middlesex University, \\
The Burroughs, London NW4 4BT, United Kingdom. }
\date{}

\maketitle

\begin{abstract}
Parameter settings in evolutionary algorithms and metaheuristics are important because such parameter values can influence the performance of algorithms under evaluation. For a given algorithm, there are many different numerical experiments to show that the algorithm can work well in practice; however, in most cases there is no theoretical analysis of parameter settings. In this work, we show that theoretical analysis using the theory of dynamical systems and evolution of population variance can give some good results in terms of parameter ranges for the bat algorithm. We also show that results from numerical experiments are consistent with theoretical bounds. Such analyses can provide good insights from different perspectives about the algorithmic characteristics such as variance evolution, transition between exploration and exploitation as well as  convergence behaviour. \\[10pt]
{\bf Citation}: Xin-She Yang and Mehmet Karamanoglu, Analysis of Parameter Settings for the Bat Algorithm Using Variance Evolution, Proceedings of International Conference on Computational Science (ICCS2026), {\it Lecture Notes in Computer Science}, vol. 16784, pp. 197-210 (2026).  \qquad [First Online: 27 June 2026 \quad \href{https://doi.org/10.1007/978-3-032-29924-6_14}{DOI} ]

\end{abstract}

\section{Introduction}

There are many algorithms and techniques for solving optimization problems in engineering and industrial applications. Most algorithms have algorithm-dependent parameters that may require a certain degree of tuning or parametric studies. As real-world design problems can be highly nonlinear and multimodal, subject to multiple constraints~\cite{Boyd2004,Yang2020Rev},
some traditional algorithms such as gradient-based algorithms cannot deal well with such nonlinear optimization problems. Nature-inspired algorithms
are a class of algorithms that can have some advantages over traditional algorithms because they use some randomization components, which makes it more likely for algorithms to find the global optimal solutions. However, most nature-inspired metaheuristic algorithms tend to have more parameters than traditional algorithms~\cite{Kennedy1995PSO,Storn1997DE,Pham2005,Yang2020Rev}, and their performance can be influenced largely by their parameter settings. Therefore, parameter tuning is important to metaheuristic algorithms.

Parameter tuning and parameter settings are important~\cite{Eiben2011,Joy2025,Hekmat2019}. There are different ways for tuning parameters in algorithms, such as the Monte Carlo based methods~\cite{Joy2025} and others~\cite{Eiben2011}.
These methods can provide some great insights into the parameter value ranges and the best parameter values for a given algorithm for solving a set of optimization problems. Apart from parameter tuning methods, there are ways for analyzing algorithms to gain insights into their parameter settings, including the use of dynamical system theory~\cite{Clerc2002} and the analysis of population variance~\cite{Zaharie2009}. In this paper, we will use the bat algorithm (BA) as the main focus to show that the evolution of population variance can obtain the same parameter ranges
as other methods. In addition, the comparison of actual variance with theoretical predictions gives constant results, and the order of transition time from exploration to exploitation for a given problem is also consistent with theoretical analysis.

This paper is organized as follows. Section 2 briefly outlines the bat algorithm and parameter settings, whereas Section 3 presents the analysis of the variance evolution in the swarm population. Section 4 uses three benchmark functions to validate and compare the variance variations during iterations. Section 5 shows the parameter settings on acceptance probability of new solutions. Finally, Section 6 provides a summary with some discussion about future work.

\section{Bat Algorithm and its Parameter Settings}

The bat algorithm (BA) is a swarm intelligence based algorithms~\cite{YangBat2010}, which has been extended and applied to many applications~\cite{YangGandomi2012BA,Gandomi2014CBA,Osaba2016BA}.
For a swarm of $n$ virtual bats at locations $x_i^t$ with
flying velocities $v_i^t$ at iteration $t$, the main
algorithmic equations of the BA can be rewritten as
\begin{align}
 \w_i & = \w_{\min}+(\w-\w_{\min}) \beta, \label{BA-eq-1} \\
 v_i^{t+1} & = \rho v_i^t +(g-x_i^t) \w_i, \label{BA-eq-2} \\
 x_i^{t+1} & = x_i^t + v_i^{t+1}, \label{BA-eq-3}
\end{align}
where $g$ is the best solution found so far at iteration $t$
among all the population $x_i^t \; (i=1,2,...,n)$.
Here, $\w_i$ is the frequency of bat $i$ in the range from
$\w_{\min}$ to $\w$ (maximum possible frequency), and $\beta$ is
a random number drawn from a uniform distribution [0, 1].
In addition, there are variations of pulse emission rate and
loudness in the standard BA~\cite{YangBat2010}.
However, we will focus on the algorithmic equations only; therefore, there are two main parameters to be considered here: inertia weight $\rho$ and
the maximum frequency range $\w$.

To analyze the characteristics and iteration behaviour, fixed point
theory based methods can usually work well for deterministic, gradient-based algorithms under certain assumptions. However, such methods do not work well
for metaheuristic algorithms because randomization and random variables are used in metaheuristic algorithms~\cite{Khamsi2001Kirk,Granas2003,Zdenek2009}.
Researchers have studied metaheuristic algorithms, such as particle swarm optimization (PSO), from different perspectives.
For example, PSO has been analyzed using simplified dynamical system framework~\cite{Clerc2002}. Genetic algorithms, differential evolution and other algorithms  have been analyzed using Markov chains and other methods to gain insights about parameters and convergence~\cite{Aytug1996,Aytug2000,Beltrami1998,Bergstra2012,Rudolph1994,Trelea2003,Chatt2006,Sudholt2010,Pan2013PSO}.

For the bat algorithm, Chen et al. used both Markov chain framework and
dynamical system~\cite{Chen2018} and obtained the appropriate
ranges of the two parameters in the BA. They rewrote the BA as a dynamical system
\be v_i^{t+1}= \rho v_i^t + (g- x_i^t) \w_i= - \w x_i^t + \rho v_i^t + \w g, \ee
\be x_i^{t+1}=x_i^t+ v_i^{t+1}=x_i^t + \rho v_i^t - \w x_i^t + \w g, \ee
which can be expressed as a matrix form
\be Y^{t+1} = A Y^t + W g, \ee
where
\be Y^t=\mat{x_i^t \\ v_i^t}, \quad A=\mat{1-\w & \rho \\ -\w & \rho}, \quad
W=\mat{\w \\ \w}. \ee
The stability theory of dynamical system requires both eigenvalues $\lambda$ of $A$ must satisfy $|\lambda| \le 1$. Their analysis~\cite{Chen2018} concludes that
\be -1 \le \rho \le 1, \quad  0 \le \w \le 2\rho +2.
\label{BA-para-100} \ee

Though such analysis is insightful, however, it does not provide direct insight into how the population in the bat algorithm evolves during iterations. A different way of analyzing the population is to see how the population variance changes over time. For example, the population variance of differential evolution (DE) has been analyzed by Zaharie~\cite{Zaharie2009,Zaharie2017}, which provides some insights about the influence of the mutation probability and variations of population variance in the DE.

For the BA, we can use a similar framework to analyze
its variance evolution. In order to carry out such analysis and
for the ease of notations for late analysis, we will drop the superscripts and use $z_i=x_i^{t+1}$, $y_i=v_i^{t+1}$, $x_i=x_i^t$, and $\w_{\min}=0$.
Thus, we can rewrite the algorithmic equations \eqref{BA-eq-2} and \eqref{BA-eq-3} as
\begin{align}
& y_i=\rho v_i+(g-x_i) \w_i, \label{BA-eq-100} \\
& z_i=x_i + y_i=x_i+\rho v_i + (g-x_i) \w_i, \label{BA-eq-200}
\end{align}
where $\w_i=\w \beta$. This set of equations will be used for analysis
in the next section.

\section{Evolution of Variance}
		
For a given random variable $X$ with a mean of $\mu$ and a variance of $\sigma^2$, the mean of an arbitrary function $\phi(X)$ can be estimated by the Taylor expansion method~\cite{Grindstead1997,Wolter1985} or the delta method in general~\cite{BenHan2005,Oehlert1992}.
\be \E{\phi(X)}=\phi(\mu) +\frac{\phi''(\mu)}{2} \sigma^2, \ee
where $\phi''$ is the second derivative of $\phi$.
The variance of $\phi(X)$ can be approximately by
\be \textrm{Var}[\phi(X)] =[\phi'(\mu)]^2 \sigma^2 + \frac{[\phi''(\mu)]^2}{2} \sigma^4. \ee

\subsection{Variance Evolution of the Bat Algorithm}

Now we can carry out the theoretical analysis of the BA using a
similar analysis for differential evolution~\cite{Zaharie2009,Zaharie2017}.
For bat $i$,  the updating equation \eqref{BA-eq-200} becomes
\be z_i =x_i+\rho v_i+(g-x_i) \w \beta, \ee
where $\beta$ obeys a uniform distribution with a mean of $\beta_0$.
When $\beta \sim U(0,1)$, we have $\beta_0=1/2$.

Similarly, the corresponding equation for bat $j$ is
\be z_j=x_j+\rho v_j +(g-x_j) \w \beta. \ee
The differences of the preceding two equations give
\be (z_i-z_j)=(x_i-x_j)+\rho (v_i-v_j) - (x_i-x_j) \w \beta, \ee
which can be written compactly as
\be (z_i-z_j)=(1-\w \beta) (x_i-x_j) + \rho (v_i-v_j). \ee
By squaring this equation, it becomes
\[ (z_i-z_j)^2=(1-\w \beta)^2 (x_i-x_j)^2 \]
\be + \rho^2 (v_i-v_j)^2 +
2 (1-\w \beta) \rho (x_i-x_j) (v_i-v_j). \ee

Though the population evolves with time, it is difficult to show if all bats or agents are completely independent in the statistical sense. However, as an approximation and for simplicity, we assume here that $(x_i-x_j)$ and $(v_i-v_j)$ are independent for a given population of size $n$. Thus, the expectation of the preceding equation becomes
\be \frac{2n}{(n-1)} \sigma_z^2 =(1-\w \beta_0)^2 \frac{2n}{(n-1)} \sigma_x^2 +\rho^2 \frac{2n}{(n-1)} \sigma_v^2. \ee
where we have used
\be \E{(x_i-x_j)^2}=\frac{2n}{(n-1)}\sigma_x^2, \ee
\be \E{(1-\w \beta) g(Z)}=\E{(1-\w \beta)} \E{g(Z)} \ee
assuming $\beta$ and $g(Z)$ are independent~\cite{BenHan2005,Zaharie2009}.

When the population size $n$ is sufficiently large ($n \gg 1$), we have approximately  $n/(n-1) \approx 1$ and
\be \sigma_z^2=(1-\w \beta_0)^2 \sigma_x^2 +\rho^2 \sigma_v^2,
\label{Var-BA-100} \ee

It should be sufficiently realistic to assume that $\sigma_v^2=\sigma_x^2$
for the previous population at an earlier iteration because the variance of velocities is closely linked to the variance of positions. Then, equation \eqref{Var-BA-100} becomes
\be \sigma_z^2 = [(1-\w \beta_0)^2 +\rho^2] \sigma_x^2. \ee
This is the change of variances in a single iteration. For $t \ge 1$ iterations, the new variance at iteration $t$ becomes
\be \sigma_{z,t}^2=\Lambda^t \sigma_{0}^2, \label{Var-eq-123} \ee
where $\sigma_0^2$ is the the variance of the initial population. The factor
is given by
\be \Lambda=(1-\w \beta_0)^2+\rho^2 \le 1. \label{Var-BA-200} \ee
The reduction of variance during the iteration requires that
\be \Lambda \le 1. \ee

At one extreme $\rho^2=1$ from $\rho=\pm 1$, it requires that
\be 1-\w \beta_0=0, \ee
or
\be \w=\frac{1}{\beta_0}=2, \ee
for a uniform distribution in [0, 1] with $\beta_0=1/2$.
At the other extreme $\rho=0$, equation \eqref{Var-BA-200} becomes
\be -1 \le 1-\w \beta_0 \le +1, \ee
which gives
\be 0 \le \w \le \frac{2}{\beta_0}=4. \ee
This conclusion is consistent with the results, given in Eq.~(\ref{BA-para-100}). That is
\be -1 \le \rho \le +1, \quad 0 \le \w \le 4. \ee
This shows that both the variance evolution and dynamical systems can obtain the same parameter value ranges for the BA.

\subsection{Transition from Exploration to Exploitation}

As the iterations and search for optimality continue, the overall
population variance should gradually be reduced. Once the variances
become sufficiently small, the diversity of the whole population
may be limited, indicating potentially converged states.

There is a transition time constant $\tau=t_*$ when the population variance reduces to about $1/100=1\%$ of its initial variance. That is
\be \frac{\sigma_{z,t}^2}{\sigma_0^2}
=\Lambda^{t_*}=[(1-\w \beta_0)^2 +\rho^2]^{t_*} =0.01.  \ee
Taking the logarithm of both sides, we have
\be \tau=t_*=\frac{\ln 0.01}{\ln[(1-\w \beta_0)^2 +\rho^2]}.  \ee
In case of $\beta_0=1/2$ and $\w=2$, we have
\be \tau \approx \left\lceil\frac{4.6}{-\ln [(1-2 \times 1/2)^2 +\rho^2]}\right\rceil
=\left\lceil\frac{2.3}{-\ln \rho} \right\rceil.  \ee
For $\rho=0.9$, we have $\tau \approx 22$ after rounding up to the nearest integer. In practice, the transition can occur between $\tau$ to $2 \tau$
(or about $22$ to $44$ iterations for $\rho=0.9$).

It is worth pointing out that the choice from the initial variance to
1/100 of the initial value is rather arbitrary.
If a different value such as 1/1000 is used,
the time constant $\tau$ only changes by a factor of $\ln 0.001/\ln 0.01
=1.5$, and this factor does not affect the results significantly.
Thus, the transition time constant should be viewed as an approximate timescale for the transition from exploration to exploitation, which can be estimated by $O(-1/\ln \rho)=O(1/\ln(1/\rho))$.

Even the variance does not change much after this transition, the objective values continue to reduce (or improve) for minimization problems because the moves are mostly local refinements. However, if the transition occurs too early, it may lead to premature convergence of the population. Therefore,
$\rho$ cannot be too small. In practice, $\rho \in [0.7, 0.97]$ is the appropriate range, as indicated from our empirical parameter studies.

\section{Numerical Experiments}

To validate the above theoretical results and compare with the actual
variations of variances for the BA, we will use three test benchmarks: the
sphere function, the Rosenbrock function and the spring design benchmark.
The sphere function is convex and separable, whereas the Rosenbrock function is non-convex and non-separable. The spring design problem is a nonlinear
constrained design problem.

\subsection{Sphere Function}

\begin{figure}
\centering
\includegraphics[height=1.5in,width=2.25in]{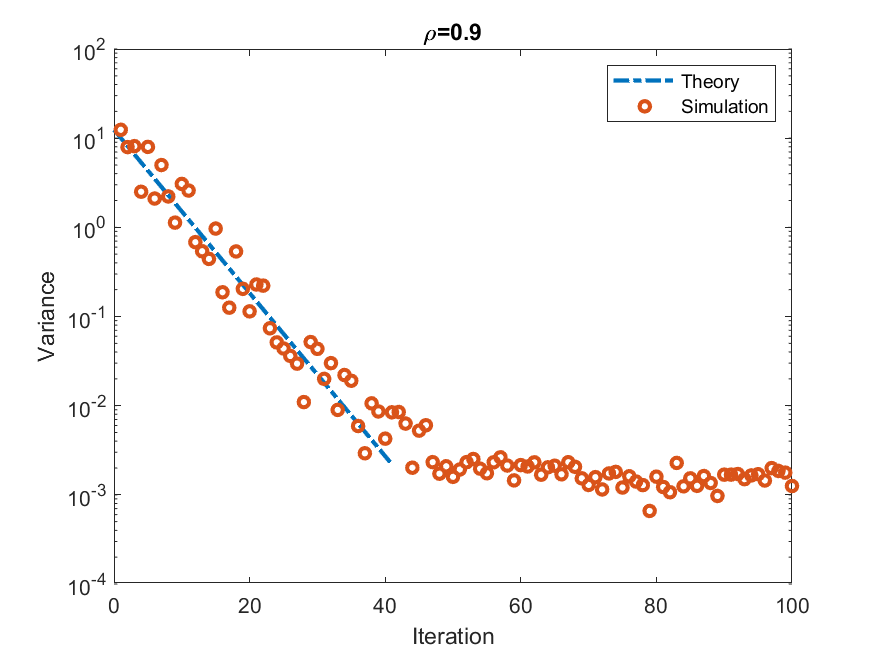}
\includegraphics[height=1.5in,width=2.25in]{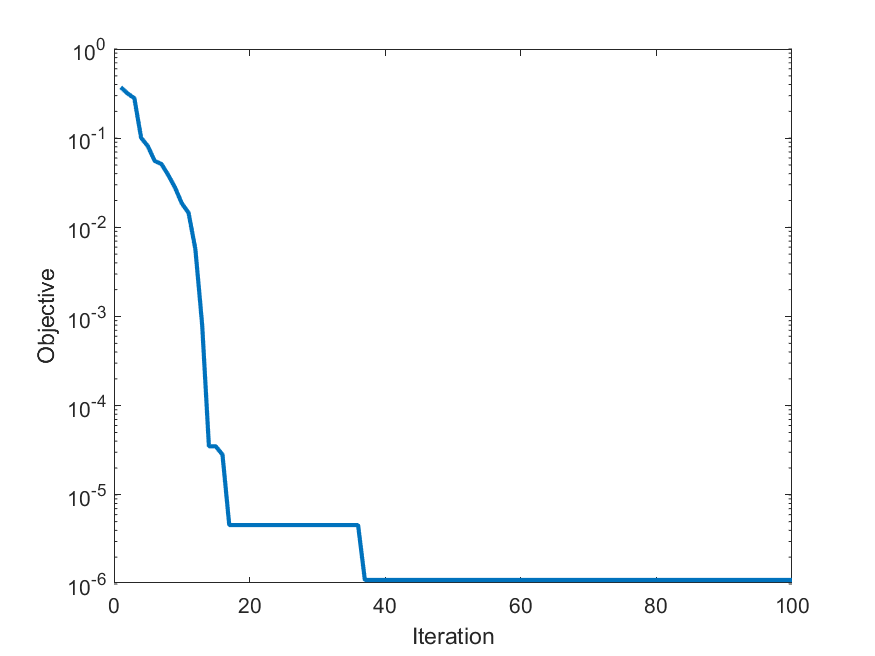}

\caption{Variation of population variance of the bat algorithm with $\rho=0.9$ for the sphere function where the dashed line is the prediction from theory and the data are obtained from simulation (left).
The variations of the objective values $f_{\min}$ are shown on the right.}
\label{Fig-var-1}
\end{figure}

For the $D$-dimensional sphere function
\be f_1(\x)=\sum_{i=1}^D x_i^2, \quad x_i \in \mathbb{R}, \ee
we can use the BA to find its minimum  $f_{\min}=0$ at $(0,0,...,0)$.
For a population of size $n=20$, the number of iteration is set to be $t_{\max}=100$. For $D=3$ and different values of $\rho=0.7, 0.8, 0.9$ with $\w=2$, the theoretical prediction of the population variances with the actual population variance in simulation are shown in Fig.~\ref{Fig-var-1}.

The transition from a quick variance reduction to a slower rate occurs between $\tau=22$ to $2 \tau=44$ iterations, as seen in Fig.~\ref{Fig-var-1}. After this transition period, even the variance does not change, the objective values continue to reduce for minimization problems, which shows that the search moves are mostly local refinements at later iterations.

\begin{figure}
\centering
\includegraphics[height=1.5in,width=2.25in]{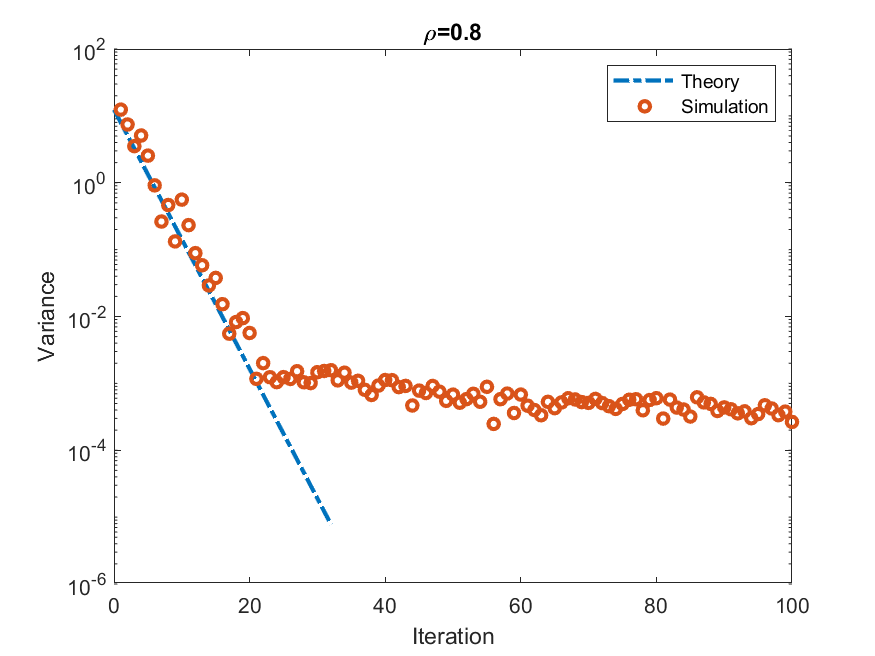}
\includegraphics[height=1.5in,width=2.25in]{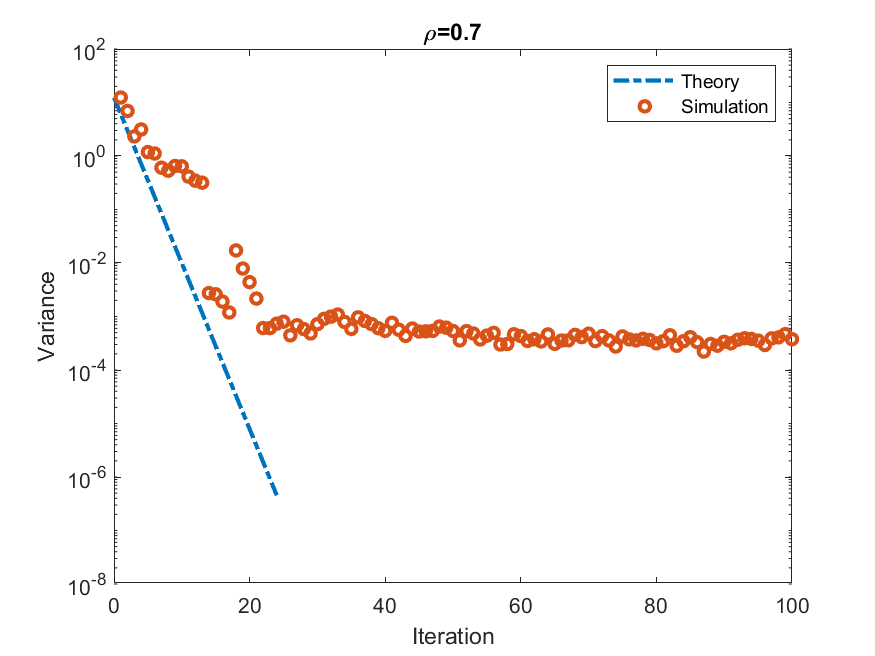}

\caption{Evolution of population variance at different stages of iterations for $\rho=0.8$ (left) and $\rho=0.7$ (right). }
\label{Fig-var-2}
\end{figure}

Similarly, variances for $\rho=0.8$ and $\rho=0.7$ are also shown in Fig.~\ref{Fig-var-2}. As we can see, the predictions are quite accurate at the earlier iterations for $\rho=0.9$ and $\rho=0.8$. When $\rho$ gets smaller, the prediction is less accurate.

In addition, at later iterations, the predictions are very different, which may indicate that the BA becomes more exploitation focused. In this case,
the search move and update of solutions are mainly local, and the acceptance rate of new moves becomes low. Thus both, the variances and the objective values become slow varying as see in the flat part of the curves.

\subsection{Rosenbrock Function}

The sphere function is a convex function, and the evolution of its variances  fits well to the theoretical trends for different values of $\rho$. Now let us test the theoretical results further using a non-convex function
\be f_2(\x)=(1-x_1)^2 + 100 \sum_{i=2}^D (x_i-x_{i-1}^2)^2, \quad
\quad -100 \le x_i \le 100. \ee

\begin{figure}
\centering
\includegraphics[height=1.5in,width=2.25in]{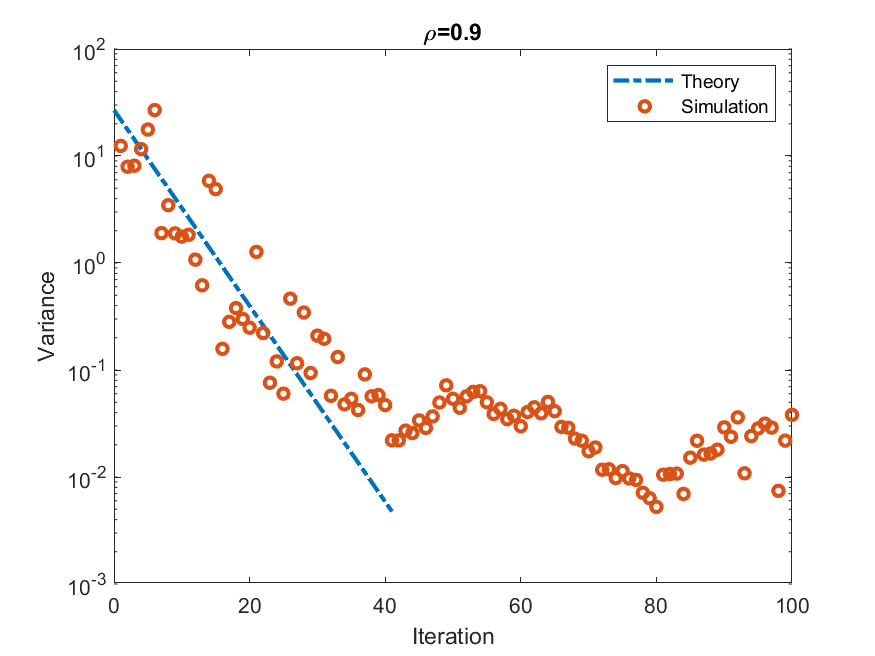}
\includegraphics[height=1.5in,width=2.25in]{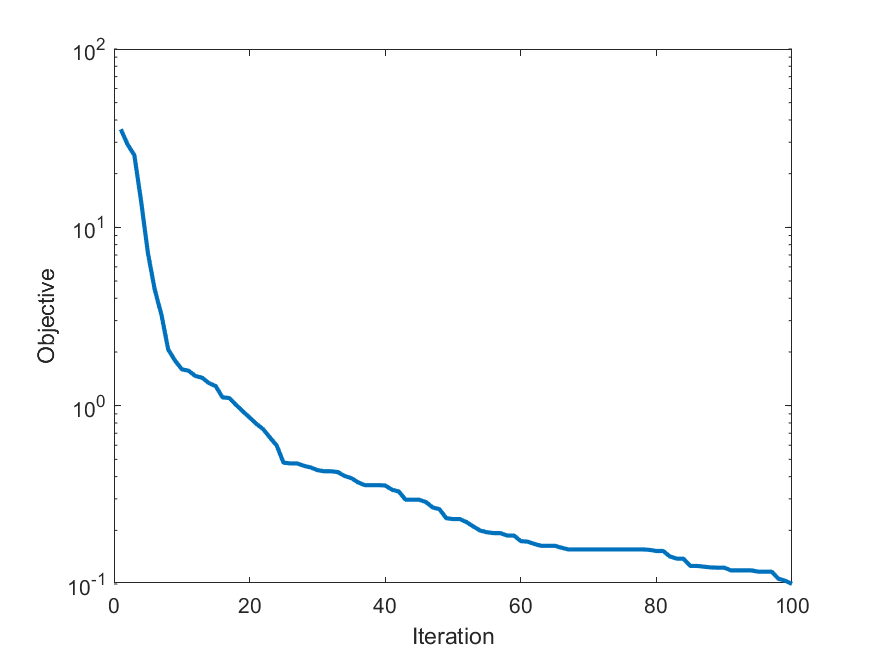}

\caption{The population variance for $\rho=0.9$ for the
Rosenbrock function (left) and the variations of objective values (right). }
\label{Fig-var-3}
\end{figure}

\begin{figure}
\centering
\includegraphics[height=1.5in,width=2.25in]{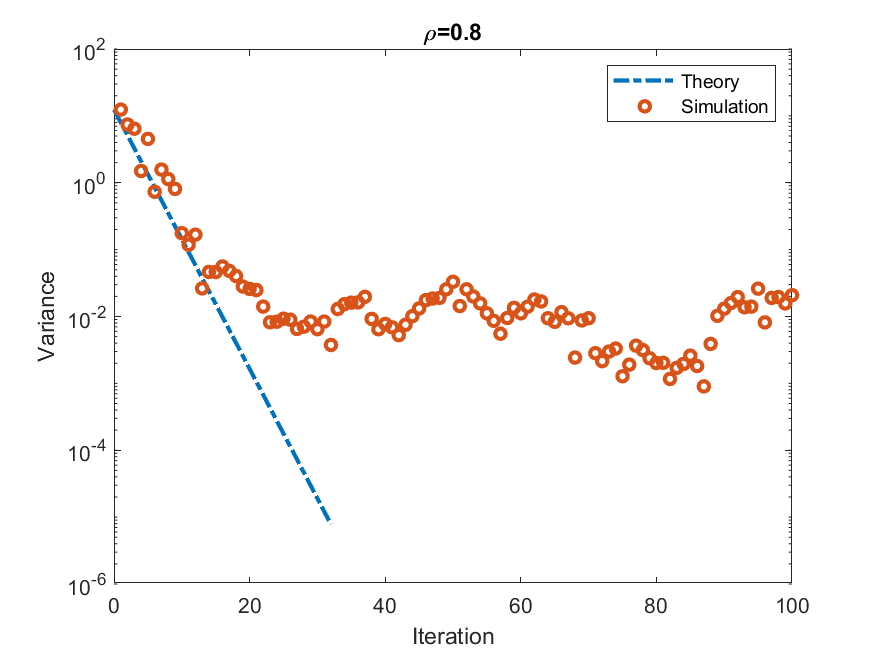}
\includegraphics[height=1.5in,width=2.25in]{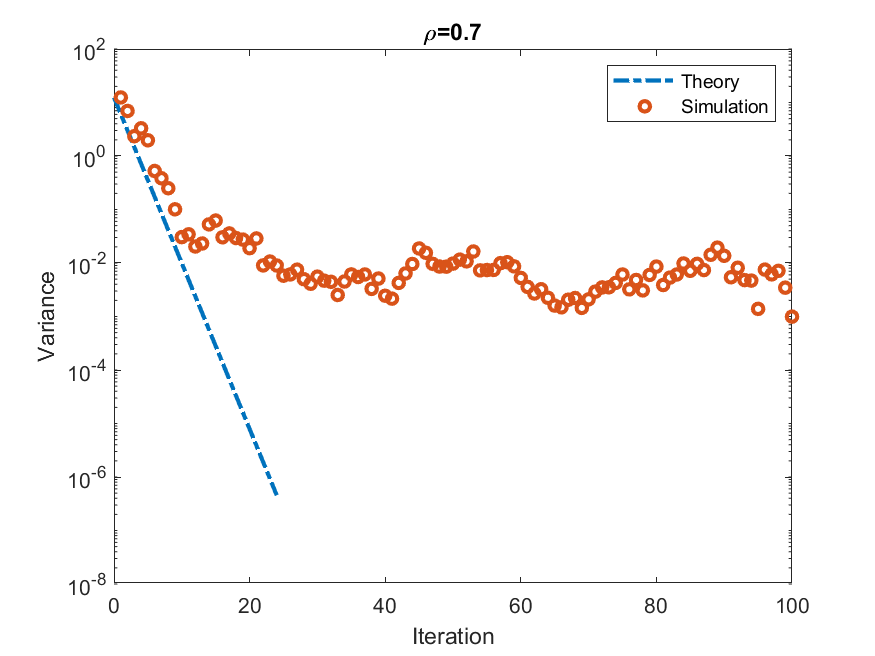}

\caption{The evolution of $f_2$ variances for $\rho=0.8$ (left) and $\rho=0.7$ (right). }
\label{Fig-var-4}
\end{figure}

Using the same parameter settings of the BA ($n=20$, $t_{\max}=100$, $D=3$, and $\rho=0.7, 0.8, 0.9$), the optimal solutions are found by the BA. The variations of the variances are shown in Figs.~\ref{Fig-var-3} and \ref{Fig-var-4} where we can see that the actual variations of variances fit reasonably well to the theoretical trend on the log scale at the initial stage of the iterations. As the iteration continues, the search process becomes more focused on the local exploitation, thus the variances becomes slow varying, but objective values continue to reduce due to local refinement.

The transition from exploration to exploitation also occurs at about 22 to 44 iterations for $\rho=0.9$. This means that the transition seems to be independent of the benchmarks in the simulation.

\subsection{Spring Design}

\begin{figure}
\centering
\includegraphics[height=1.5in,width=2.25in]{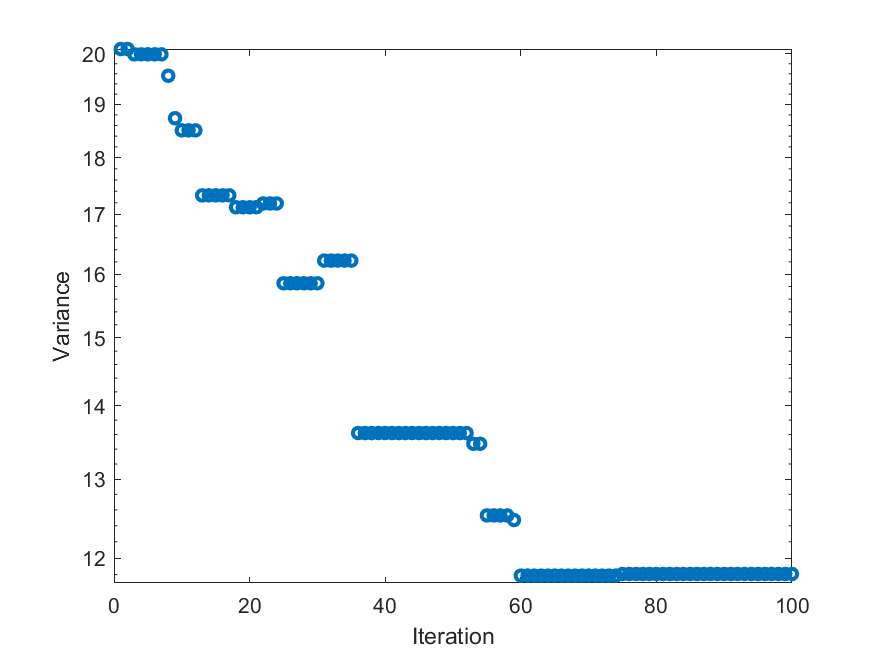}
\includegraphics[height=1.5in,width=2.25in]{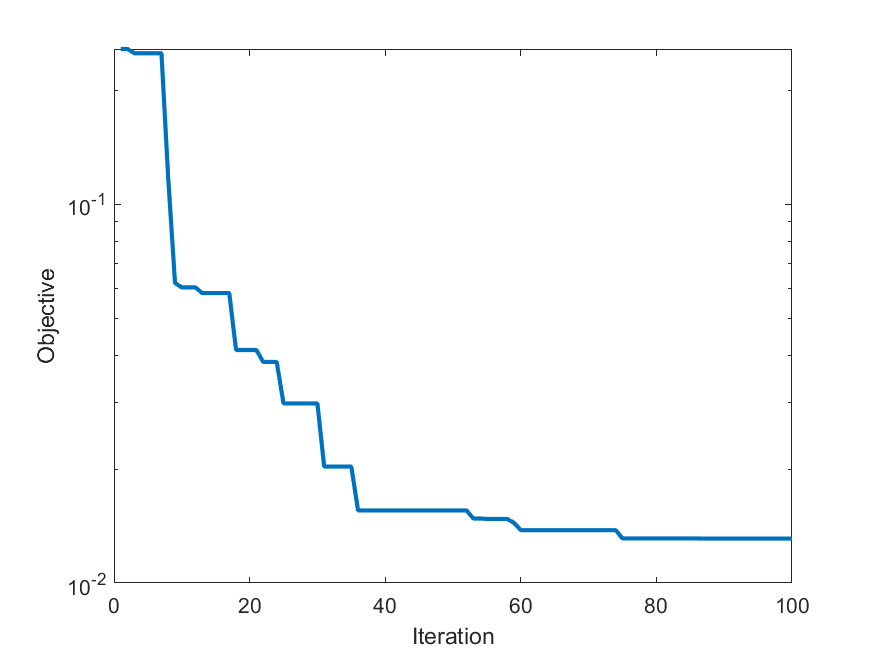}

\caption{The evolution of variances and spring design objective values for $\rho=0.7$. }
\label{Fig-var-5}
\end{figure}

The spring design problem is to manufacture a compression and tension spring from a wire of diameter $w$, which will form a spring with a mean coil diameter $D$ and the number $m$ of active coils. The objective is to minimize the overall weight or mass
\be \min \; f(w,D,m)=(2+m) w^2 D, \ee
subject to four constraints
\begin{align}
g_1 & = 1-\frac{m D^3}{71785 w^4} \le 0, \\
g_2 & = \frac{(4 D^2-w D)}{12566 (D w^3-w^4)} + \frac{1}{5108 w^2}-1 \le 0, \\
g_3 & = 1-\frac{140.45 w}{m D^2} \le 0, \\
g_4 & = \frac{(w + D)}{1.5} -1 \le 0. \\
\end{align}
These constraints are related to deflection, shear stress, frequency and physical dimension constraints. In the simple domain of
\be 0.05 \le w \le 1.00, \quad 0.25 \le D \le 1.30, \quad 2 \le m \le 15, \ee
the best solution in the current literature~\cite{Cag2008,GandomiYang2013BA} is
\be f_{\min}=0.01266522, \quad w=0.05169, \quad D=0.35673,
\quad m=11.2885. \ee
With the same parameter settings ($t_{\max}=100$, $n=20$, $\rho=0.7$), we have run the BA to find its optimal solutions. The obtained objective values and variances are plotted in Fig.~\ref{Fig-var-5}.

From Fig.~\ref{Fig-var-5}, we can see that the variances reduce quite rapidly at the initial stage of iterations, and the transition from exploration
to exploitation seems to occur between 35 to 60 iterations. Though the variances do not change much from iteration 60 onwards, the objective continues to improve/reduce gradually, which means that late iterations are mainly local exploitative moves. This characteristic is
consistent with the theoretical results.

\section{Probability of Acceptance}

In the above analysis, we have considered only two parameters $\rho$ and $\w$. In the BA, there is a switching probability between two branches based on the decision about the new solution. If a new solution is better, it will be accepted. If the solution is partially
improved (i.e., not the new best solution but better than some of the old solutions in the population), it will be accepted by comparing with a random number. In essence, there is a probability $p$ of acceptance for a given new solution.

In the rest of this paper, let us explore the possible influence of parameter settings on the effective acceptability $p$ of new moves or solutions. For a new search move or a new solution, it will be accepted with a probability $p$ and rejected with a probability of $1-p$. Alternatively, the solution in the new iteration will be the new solution with probability of $p$, while keeping the old solution $x_i$ with a probability of $1-p$. Thus, equation~\eqref{BA-eq-200} can be rewritten as
\[ z_i= (1-p) \underbrace{x_i}_{{\rm old}}
+ p \underbrace{[x_i +\rho v_i+ (g-x_i) \w_i]}_{{\rm new \; solution}} \]
\be = (1-p) x_i + p x_i + \rho p v_i + p g \w \beta-p \w \beta x_i,
\label{BA-var-250} \ee
where we have used $\w_i=\w \beta$ with $\beta$ being a uniform distribution.

In our derivations, we will need to use the properties of the variance of two uncorrelated random variables $U$ and $V$ with means $\mu_U$ and $\mu_V$ and variances $\sigma_U^2$ and $\sigma_V^2$, respectively.  The variance of their product can be calculated by
\be \sigma_{UV}^2=\textrm{Var}(U V)=(\sigma_U^2+\mu_U^2)(\sigma_V^2+\mu_V^2)-\mu_U^2 \mu_V^2, \ee
with the standard properties
\[ \textrm{Var}(a U \pm b V)=a^2 \sigma_U^2+b^2 \sigma_V^2 \pm 2 ab \; \textrm{cov}(U,V)\]
\be = a^2 \sigma_U^2+b^2 \sigma_V^2, \quad a, b \in \mathbb{R},
\label{Var-100} \ee
where we have assumed that $U$ and $V$ are uncorrelated. That is, cov($U,V$)=0 in the current context for the swarm population.
Again as pointed out earlier, this assumption is an approximation.

Let $\sigma_x^2$ be the variance of $x_i$, $\sigma_v^2$ be the variance of $v_i$, and $\sigma_z^2$ be the variance of $z_i$.
By taking the variance of Eq.~\eqref{BA-var-250}, we have
\[ \sigma_z^2=(1-p)^2 \sigma_x^2 + p^2 \sigma_x^2 \]
\be + p^2 \rho^2 \sigma_v^2 + p^2 \w^2 \textrm{Var}(g \beta) + \w^2 p^2 \textrm{Var}(x_i \beta). \label{Var-200} \ee
Here, we can assume that $g$ and $\beta$ are uncorrelated, and
$x_i$ and $\beta$ are also uncorrelated.

The mean and variance of a uniform distribution $\beta \in [0, 1]$ are
$\mu_{\beta}=1/2$ and $\sigma_{\beta}^2=1/12$, respectively. The mean of $g$ is $g$, and the variance of $g$ is $\sigma_g^2$, which can be considered
as the same as $\sigma_x^2$. Similarly, the mean and variance of $x_i$ are
$\mu_x$ and $\sigma_x^2$, respectively. As the current best solution
is usually obtained from $x_i$, we can approximate it as $g=\mu_x$.
Thus, we have
\be \textrm{Var}(g \beta)=(\sigma_x^2+g^2)\left[\frac{1}{12}+\left(\frac{1}{2}\right)^2 \right] -g^2 \left(\frac{1}{2}\right)^2=\frac{\sigma_x^2}{12} + \frac{g^2}{12} +\frac{\sigma_x^2}{4}, \ee
and
\[ \textrm{Var}(x_i \beta)=(\sigma_x^2 + \mu_x^2) \left[\frac{1}{12}+\left(\frac{1}{2}\right)^2 \right]-\mu_x^2 \left(\frac{1}{2}\right)^2  \]
\be = \frac{\sigma_x^2}{12}+\frac{g^2}{12}+\frac{\sigma_x^2}{4}. \ee

Now Eq.~\eqref{Var-200} becomes
\be \sigma_z^2=(1-p)^2 \sigma_x^2 +p^2 \sigma_x^2 + p^2 \rho^2 \sigma_v^2
+p^2 \w^2 \left[\frac{2 \sigma_x^2}{3} + \frac{g^2}{12} \right]. \ee

For simplicity, we can also assume that $\sigma_v^2$ is the same as that for $x_i$ in the previous iteration (from Eq.~\eqref{BA-eq-3}).

With the above assumptions and notations, the preceding equation becomes
\[ \sigma_z^2=\left[(1-p)^2 + \frac{2 p^2 \w^2}{3} + p^2 \right] \sigma_x^2 + p^2 \rho^2 \sigma_x^2 \]
\be =\left[(1-2p+2p^2 +\frac{2 p^2 \w^2}{3} +p^2 \rho^2\right] \sigma_x^2 + p^2 \w^2 \frac{g^2}{12}. \ee
For many test functions or design benchmarks,
the best solutions are either close to $x_*=0$ or a known vector constant. Thus, we can consider the final term on the right-hand side of the above equation as a constant or zero in most cases.

Thus, the reduction of the variance and the convergence of the algorithm requires that
\be Q=1-2p + 2p^2 +\frac{2 p^2 \w^2}{3} + p^2 \rho^2 \le 1, \ee
which means that
\be p \le \frac{2}{2 + 2 \w^2/3 +\rho^2}
=\frac{1}{1 +\w^2/3+\rho^2/2}, \ee
where we have used the fact that $p>0$. It is easy to see that this result implies that  $0<p \le 1$ is always true.

For the special case of $\rho=1$ and $\w=4$, we have
\be \rho \le \frac{6}{41} \approx 0.146. \ee
This is the maximum theoretical probability of accepting new solutions,
but the true rate of acceptance of new moves can be lower in practice.
The comparison of actual probability of acceptance is summarized in Table~\ref{Tab-rate-1}. As we can see, the theoretical values provide an
upper bound for the true acceptance probability. Obviously, the true
acceptance rate can depend on the ways of generating new solutions,
the diversity of population, feasibility of the new solutions and the
locality of the new solutions.

\begin{table}
  \centering
 \caption{Comparison of the actual acceptance probability with the theoretical limit.}
  \begin{tabular}{|c|c|c|c|} \hline
 Function &  $\rho$ & Maximum probability &  Actual \\
         &          & from theory  &         Rate \\ \hline
 $f_1$ & 0.9 & 0.365 & 0.140 \\
      & 0.8 & 0.377 &  0.102 \\
      & 0.7 & 0.388 & 0.091 \\ \hline
 $f_2$ & 0.9 & 0.365 & 0.307 \\
       & 0.8 & 0.377 & 0.224 \\
      & 0.7 & 0.388 & 0.258 \\
     \hline

 Spring & 0.9 & 0.365 & 0.076 \\
        & 0.8 & 0.377 & 0.089 \\
        & 0.7 & 0.388 & 0.115  \\ \hline
   \end{tabular}
 \label{Tab-rate-1}
\end{table}

\section{Conclusion and Discussion}

We have analyzed the evolution of variance for the bat algorithm in this work and have shown that the parameter ranges are consistent with the results from other theoretical analysis using dynamical system theory. In addition, the change in the variance reduction can also indicate a transition in the search mechanism from global exploration to local exploitation. The characteristics of variance evolution can be predicted well at the earlier stage of the iterations. This shows that the theoretical insights and results are correct.

We also have estimated the upper limit of the acceptance probability of new solutions in the population using variance analysis,
and the results from numerical experiments are largely consistent with theoretical indications. The difference between the theoretical bounds and the actual acceptance rates means that the theory tends to over-estimate the probability. Thus, there is room for improvement to provide a tight bound. This can form a research topic for further studies.

Furthermore, the current work indicates that the transition timescale seems to be independent of the problems to be solved. In practice, there may be other factors that could affect the behaviour of a swarm population, such as population size, modality of the problem, and nonlinearity of the constraints as well as the initialization of the swarm population. This work can form the basis for further research.

\subsubsection{Declaration}

The authors confirm that there are no relevant financial or non-financial competing interests to report.

No funding was received for this research work.

No datasets were received/used. All results are simulated and reproducible
from the proposed method.

\end{document}